\documentclass[nohyperref]{article}

\usepackage{microtype}
\usepackage{graphicx}
\usepackage{caption}
\usepackage{subcaption}
\usepackage{booktabs} 

\usepackage{hyperref}

\usepackage[accepted]{icml2022}

\usepackage{amsmath}
\usepackage{amssymb}
\usepackage{mathtools}
\usepackage{xspace}
\usepackage{amsthm}

\usepackage{amsmath,amsfonts,bm}
\usepackage{amsthm}
\usepackage{thmtools,thm-restate}

\def\Figref#1{Figure~\ref{#1}}

\def\Secref#1{Section~\ref{#1}}

\def\eqref#1{Equation~\ref{#1}}

\def\1{\bm{1}}

\DeclareMathAlphabet{\mathsfit}{\encodingdefault}{\sfdefault}{m}{sl}
\SetMathAlphabet{\mathsfit}{bold}{\encodingdefault}{\sfdefault}{bx}{n}

\usepackage{algorithm}
\usepackage[noend]{algpseudocode}
\usepackage{multirow}

\usepackage[capitalize,noabbrev]{cleveref}

\usepackage[textsize=tiny]{todonotes}

\newcommand{\tabref}[1]{Table~\ref{#1}}

\newcommand{\ie}{\emph{i.e.},\xspace}
\newcommand{\etc}{etc.\xspace}

\newcommand{\supervised}{Sup\xspace}
\newcommand{\supcon}{SCL\xspace}
\newcommand{\oneshot}{One-Shot\xspace}
\newcommand{\gmp}{GMP\xspace}
\newcommand{\gmpdelayed}{$\Delta$GMP\xspace}
\newcommand{\qscore}{Q-Score\xspace}
\newcommand{\zscore}{Z-Score\xspace}
\newcommand{\pdepth}{PD-Score\xspace}

\renewcommand{\algorithmicrequire}{\textbf{Input:}}
\renewcommand{\algorithmicensure}{\textbf{Output:}}

\icmltitlerunning{Studying the Impact of Magnitude Pruning on Contrastive Learning Methods}
 
\begin{document}

\twocolumn[
\icmltitle{Studying the impact of magnitude pruning on contrastive learning methods}



\icmlsetsymbol{equal}{*}

\begin{icmlauthorlist}
\icmlauthor{Francesco Corti}{equal,yyy}
\icmlauthor{Rahim Entezari}{equal,comp}
\icmlauthor{Sara Hooker}{cr}
\icmlauthor{Davide Bacciu}{yyy}
\icmlauthor{Olga Saukh}{comp}
\end{icmlauthorlist}

\icmlaffiliation{yyy}{University of Pisa}
\icmlaffiliation{comp}{TU Graz / CSH Vienna}
\icmlaffiliation{cr}{Cohere.ai}


\icmlkeywords{Machine Learning, ICML}

\vskip 0.3in
]



\printAffiliationsAndNotice{\icmlEqualContribution} 


\begin{abstract}
We study the impact of different pruning techniques on the representation learned by deep neural networks trained with contrastive loss functions. Our work finds that at high sparsity levels, contrastive learning results in a higher number of misclassified examples relative to models trained with traditional cross-entropy loss. To understand this pronounced difference, we use metrics such as the number of PIEs~\cite{hooker2019compressed}, \qscore~\cite{kalibhat2022understanding} and \pdepth~\cite{baldock2021deep} to measure the impact of pruning on the learned representation quality. Our analysis suggests the schedule of the pruning method implementation matters. We find that the negative impact of sparsity on the quality of the learned representation is the highest when pruning is introduced early-on in training phase.
\end{abstract}


\section{Introduction}
\label{sec:intro}

Neural network pruning methods have demonstrated that it is possible to achieve high network compression levels with surprisingly little degradation of generalization performance~\citep{han2015deep, han2015learning}. On the one hand, different classes are disproportionately impacted by network pruning, and some examples seem to be easier to forget than others~\citep{hooker2019compressed, entezari2019class}. These examples include data points in the long-tailed part of the distribution, examples that are incorrectly or imprecisely labeled, examples that have multiple labels, \etc. \citet{timpl2021understanding} show that if we keep the network capacity fixed by increasing network width and depth, pruned networks consistently match and often outperform their initially uncompressed versions. Recent work suggests the generalization properties of this setting may differ in important ways, for example ~\citep{arxiv.2010.08127} shows that contrastive representation learning, can achieve outstanding accuracy and improve out-of-distribution generalization. \emph{Are the same examples also lost if network compression is applied in a semi-supervised training regime?}



\setlength{\textfloatsep}{8pt}
\begin{figure}[t]
    \centering
    \begin{subfigure}[b]{\linewidth}
        \includegraphics[width=0.33\linewidth]{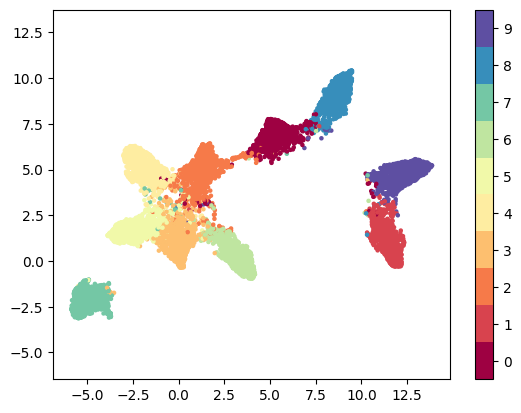}
        \includegraphics[width=0.32\linewidth]{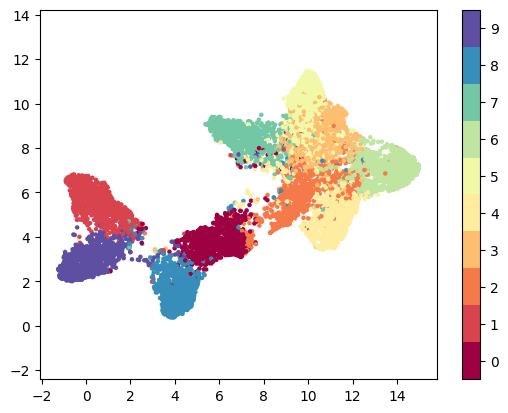}    
        \includegraphics[width=0.32\linewidth]{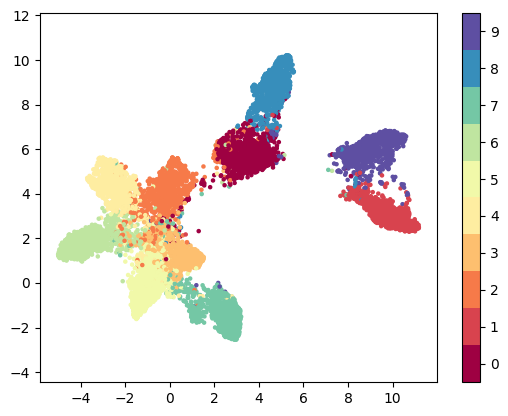}
        \caption{Representation by \emph{supervised} training without pruning (\textbf{left}), pruned 95\% by \gmp (\textbf{center}) and \oneshot (\textbf{right}).}
    \end{subfigure}
    \hfill
    \begin{subfigure}[b]{\linewidth}
        \includegraphics[width=0.325\linewidth]{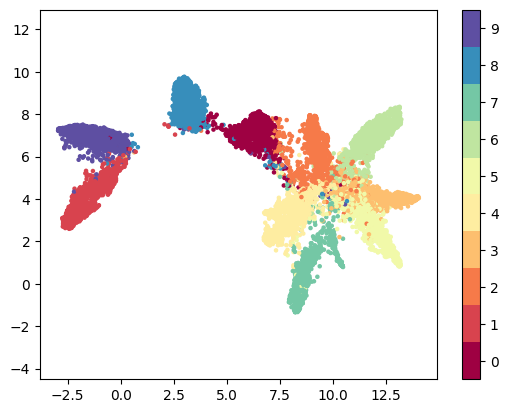}
        \includegraphics[width=0.325\linewidth]{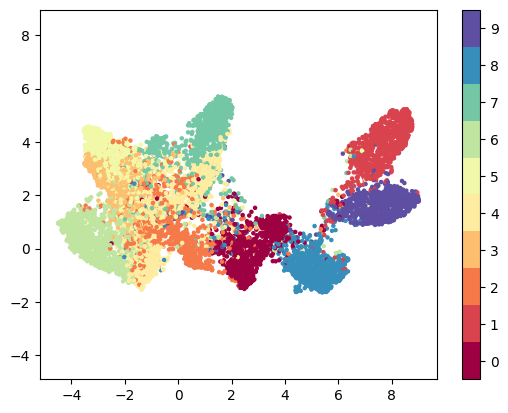} 
        \includegraphics[width=0.325\linewidth]{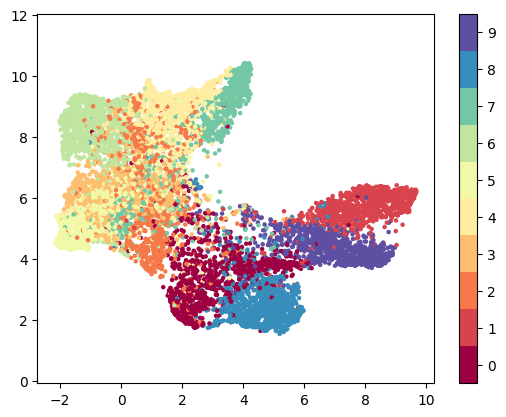} 
        \caption{Representation by \emph{supervised contrastive} training without pruning (\textbf{left}), pruned 95\% by \gmp (\textbf{center}) and \oneshot (\textbf{right}).}
    \end{subfigure}
    \caption{\textbf{Impact of pruning on the learned representation.} UMAP diagrams of the models trained with supervised (\textbf{top row}) and supervised contrastive (\textbf{bottom row}) learning using WideResNet on CIFAR-10. 
    Supervised contrastive learning is more susceptible to representation quality reduction at high sparsity than supervised learning.} 
    \label{tab:firstfigure}
\end{figure}

This paper studies the impact of different versions of magnitude pruning on the representation learned by deep models trained with a supervised cross-entropy loss (\supervised) and supervised contrastive learning (\supcon)~\cite{khosla2020supervised} methods. We investigate the impact of changing the pruning schedule, comparing the impact of post-training global one-shot pruning (\oneshot)~\cite{paganini2020streamlining}, Gradual Magnitude Pruning (\gmp)~\cite{zhu2017prune} and a delayed version of \gmp (\gmpdelayed) on the learned representation. Along with the obtained test set accuracy and the analysis of Pruning Identified Exemplars (PIEs)~\citep{hooker2019compressed}, we also evaluate the \qscore~\cite{kalibhat2022understanding}, prediction depth \pdepth~\cite{baldock2021deep} metrics and visually inspect obtained representations with UMAP~\cite{McInnes2018} to draw our conclusions. 

This work makes the following contributions:
\begin{itemize}
    \item We show that models trained with \supcon are significantly more impacted by all considered sparsification methods than the models trained with supervised learning trained to the same levels of sparsity. The negative impact is the highest early-on in training.
    \item We found that with higher sparsity representations learned by \supcon decline more in quality than for supervised learning. This motivates the need for pruning methods that are more friendly to \supcon models.
\end{itemize}

The next section details our experimental setup and the measures we use to assess representation quality. To quantify the amount of examples lost due to pruning, we use the evaluation framework introduced by \citet{hooker2019compressed} and find Pruning Identified Exemplars (PIEs) that are systematically more impacted by the introduction of sparsity. We then compare the values of several representation quality measures found in the literature to support our hypothesis that contrastive methods suffer more from sparsification than supervised training, in particular in the early phases of training.

\begin{table*}[t]
\centering
	\begin{tabular}{c|cccc|cccccc}
    \toprule
    \textbf{Spar-} &
    \multicolumn{2}{c}{ \textbf{\supervised~/~\gmp}} & 
    \multicolumn{2}{c}{ \multirow{2}{*}{\textbf{\supervised~/~\oneshot}}} & 
    \multicolumn{2}{c}{ \multirow{2}{*}{\textbf{\supcon~/~\gmp}}} & 
    \multicolumn{2}{c}{ \multirow{2}{*}{\textbf{\supcon~/~\gmpdelayed}}} & 
    \multicolumn{2}{c}{ \multirow{2}{*}{\textbf{\supcon~/~\oneshot}}} \\
    \textbf{sity} &
    \multicolumn{2}{c}{\textbf{(Ours)}} &
    & &
    & &
    & &
    & \\
    \% &
    PIE{$^\star$} & Acc[\%] &
    PIE & Acc[\%] &
    PIE & Acc[\%] &
    PIE & Acc[\%] &
    PIE & Acc[\%] \\
    \midrule
    0   & -   & 90.58$\pm 0.38$   & -   &  90.58$\pm 0.38$     & -   &  92.06 $\pm 0.17$    & -   &  92.06 $\pm 0.17$      & -   & 92.06$\pm 0.17$ \\
    30   & 188 & 90.5$\pm 0.39$  & \textbf{90}  & 90.3$\pm 0.39$ & 180 & 91.58$\pm 0.2$ & \textbf{176} & 91.59$\pm 0.18$ & 233 & 93.28$\pm 0.17$  \\
    50   & 179 & 90.4$\pm 0.34$ & \textbf{95}  & 90.37$\pm 0.37$ & 241 & 90.92$\pm 0.18$ & \textbf{184} & 91.23$\pm 0.16$ & 236 & 93.06$\pm 0.2$ \\
    70   & 204 & 90.14$\pm 0.34$ & \textbf{97}  & 90.33$\pm 0.37$ & 350 & 89.58$\pm 0.2$ & 271 & 90.48$\pm 0.25$ & \textbf{235} & 92.38$\pm 0.17$ \\
    90   & 273 & 89.47$\pm 0.36$ & \textbf{183} & 89.64$\pm 0.34$ & 748 & 85.56$\pm 0.4$ & \textbf{522}$^\diamond$  & 87.76$\pm 0.33^\diamond$ & 549 & 87.83$\pm 0.25$ \\
    \bottomrule
    \end{tabular}
    \caption{\textbf{Comparison of the number of PIEs for different training and pruning methods.} \supcon models show a more pronounced drop in performance (higher number of PIEs, lower accuracy) with higher sparsity compared to \supervised models. Pruning later in training (\oneshot pruning and \gmpdelayed) is more friendly to representation learning than early sparsification in training (\gmp). $^\star$Our results differ from those reported in \cite{hooker2019compressed} due to a different model architecture. $^\diamond$Results are aggregated over 25 models instead of 30.} 
    \label{tab:PIEsComparison}
\end{table*}

\section{Training, Sparsification and Measures}
\label{sec:measures}


\paragraph{Training methods.}
We use a WideResNet~\cite{zagoruyko2016wide} architecture trained on CIFAR-10~\cite{cifar100} throughout our experiments for \supcon and \supervised training. The encoder part of the network contains 690k parameters and it is followed by a classification head realizing part of the supervised task. For \supcon training our setup follows the structure proposed in~\cite{chen2020simple}. The encoder is first trained with a contrastive loss on an augmented version of the dataset by using a projection head discarded at the end of training, then the classification head is fine-tuned with supervised training. We use the code from the official repository~\cite{HobbitLong} and follow the parameter settings used in~\cite{hooker2019compressed} and summarized in Appendix~\ref{sec:training_params}. Pytorch code to reproduce the experiment is released at \href{https://anonymous.4open.science/r/What-Is-Lost-in-Compressed-Models-Trained-with-Supervised-Contrastive-Learning-D4EF/README.md}{anonymous.4open.science/What-Is-Lost}. 

\paragraph{Pruning methods and schedules.}
We implement different versions of magnitude pruning to compare the relative merits and impact of different techniques. Global post-training one-shot pruning (\oneshot)~\cite{paganini2020streamlining} is the simplest method operating on a fully learned representation. We use fine-tuning to recover accuracy loss due to \oneshot pruning. Fine-tuning is performed using a Supervised loss function. Gradual Magnitude Pruning (\gmp)~\cite{zhu2017prune} is also used in \cite{hooker2019compressed} to investigate how pruning impacts different classes. A delayed version of \gmp (\gmpdelayed) starting from epoch 50 is evaluated for contrastive models. The \gmpdelayed pruning is used to compare earlier pruning against later pruning in terms of the impact on the quality of the compressed representation. 

\paragraph{PIEs.} We evaluate all methods based on their generalization performance, \ie test-set accuracy. To quantify the impact of compression on the model performance, we assess the level of disagreement between the predictions of compressed and non-compressed networks on a given example. As defined in \citep{hooker2019compressed}, for each sample image $i$ and a set of trained models M, we denote the class predicted most frequently by the $t$-pruned models by $y_{i,t}^{M}$ and the class predicted most frequently by uncompressed models ($t=0$) by $y_{i,0}^{M}$. The sample $i$ is classified as a Pruning Identified Exemplar (PIE) iff the label is different between the set of $t$-pruned models and the set of dense models.
\begin{equation}
    \textrm{PIE}_{i,t}= \begin{cases}
    1 &\textrm{if }  y_{i,0}^M \neq y_{i,t}^M \\
    0 &\textrm{otherwise}.
    \end{cases}
\end{equation}
As reported in \citep{hooker2019compressed} pruning introduces selective forgetting, \ie the impact of sparsity is not equally distributed amongst classes. Instead, some classes are more impacted than others. This unbalanced forgetting phenomena is stronger with lower sparsity and tends to be amplified as sparsity increases. The PIEs framework helps to quantify the impact of sparsity on the classes affected by pruning by detecting the examples for which the average prediction changes due to model pruning compared to the uncompressed version of the same model.

\begin{algorithm}[t]
\begin{algorithmic}
\Require Batch size $N$, trained encoder $f(\cdot)$, test set $D$;
\State Load $D$ with batch size $N$. Representation vector $h_{i}$ is flattened to obtain features vector of the last convolutional layer of the encoder network $f(s_{i})$. 
\Ensure \{\qscore, \zscore and $\| h_{i} \|_{1}$\}.
    \For{minibatch $\{x_{i}\}_{i=1}^{N}$}
        \For{i $\in \{1,\dots, N\}$}
            \State $h_{i} \gets f(s_{i}) \in \mathbb{R}^l$  
            \State $h_{i} \gets \frac{h_{i}}{\|h_{i}\|_{2}}$
            \Comment{Perform L2 normalization}
            \State $\text{Q}_{i} \gets \frac{\text{Z-Score}(h_{i})}{\| h_{i} \|_{1}}$
            \Comment{Compute \qscore}
            \State Store $\| h_{i} \|_{1}$, \zscore and \qscore for analysis
        \EndFor
    \EndFor    
\end{algorithmic}
\caption{\qscore computation procedure}
\label{alg:subspace}
\end{algorithm}


\setlength{\textfloatsep}{4pt}
\begin{figure*}[t]
    \centering
    \includegraphics[width=0.245\linewidth]{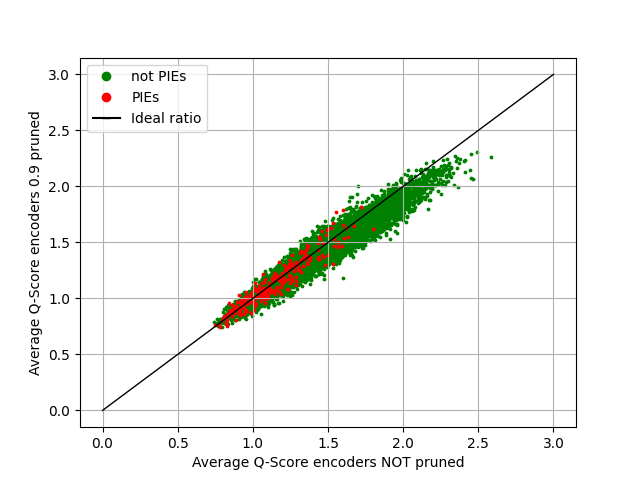}
    \includegraphics[width=0.245\linewidth]{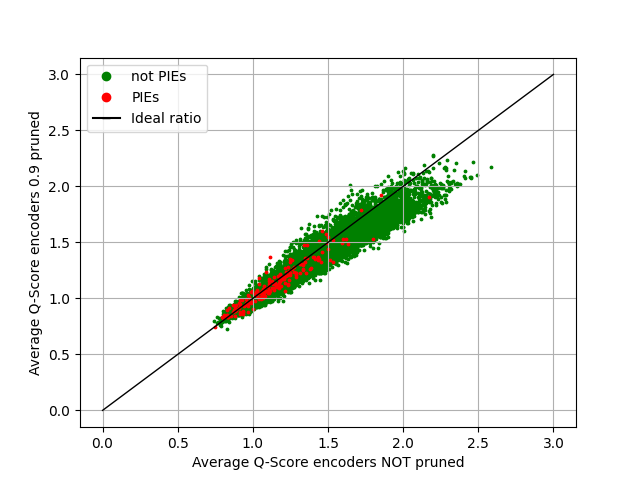}
    \includegraphics[width=0.245\linewidth]{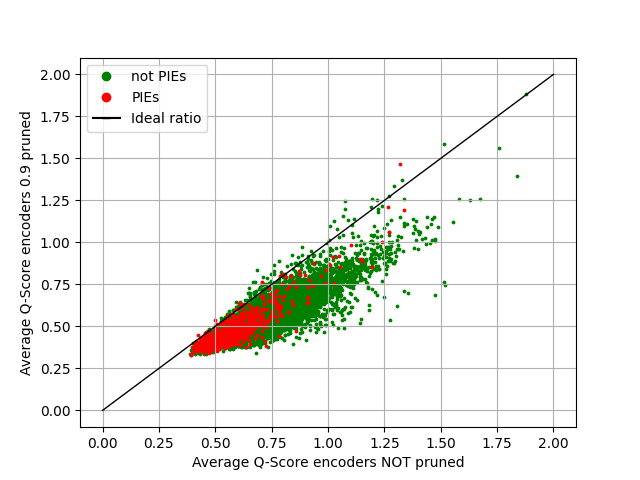}
    \includegraphics[width=0.245\linewidth]{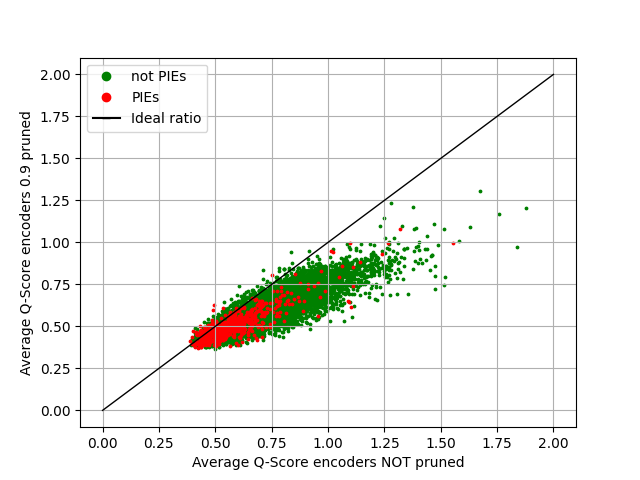}
    \caption{\textbf{\qscore analysis of the impact of pruning on the quality of learned representation.} In all plots \qscore of a non-pruned network (x-axis) is compared to the \qscore of a pruned network with 90\% sparsity (y-axis). \textbf{Left to right}: \supervised~/~\gmp, \supervised~/~\oneshot, \supcon~/~\gmp, \supcon~/~\oneshot. \qscore of \supcon models degrades at high sparsity while for \supervised models it remains stable. See Appendix \Secref{sec:scores}, \tabref{tab:qscore} and \tabref{tab:qscore-appendix} for the \qscore values for other sparsity levels.}
    \label{fig:qscore-appendix}
\end{figure*}

\setlength{\textfloatsep}{4pt}
\begin{figure*}[t]
    \centering
    \includegraphics[width=0.245\linewidth]{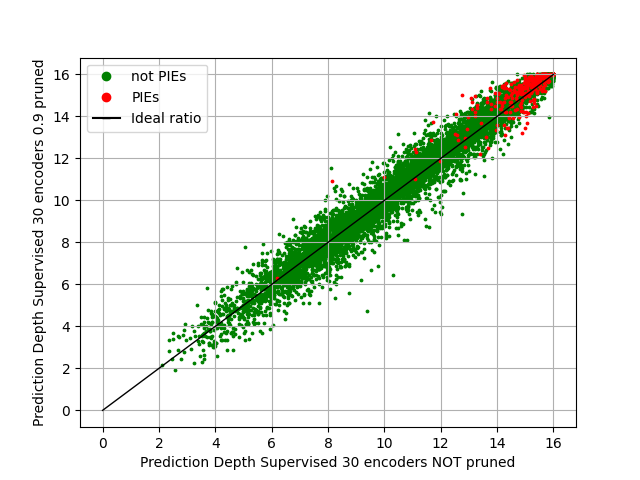}
    \includegraphics[width=0.245\linewidth]{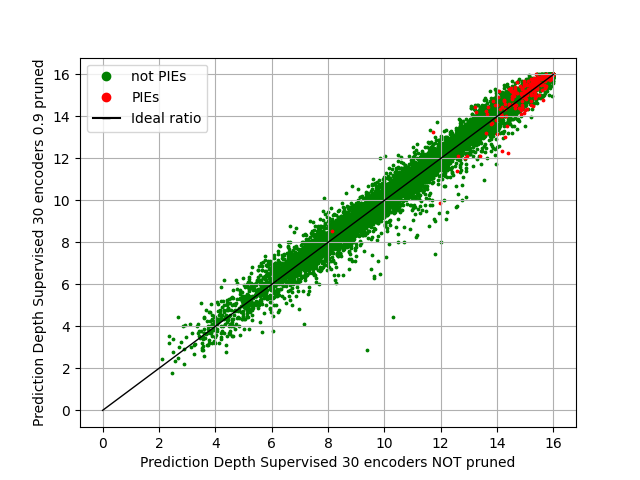}
    \includegraphics[width=0.245\linewidth]{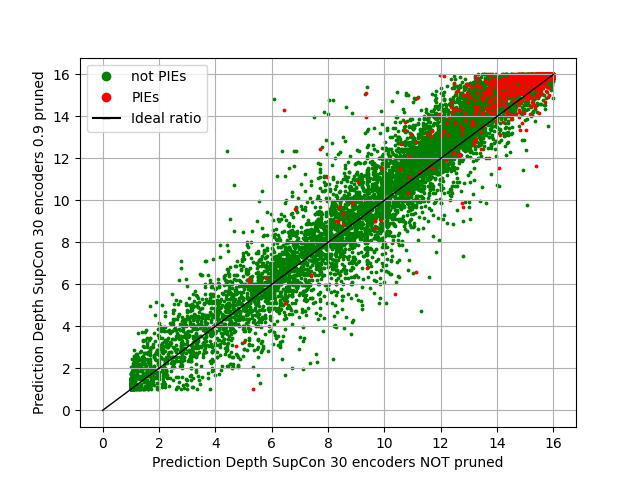}
    \includegraphics[width=0.245\linewidth]{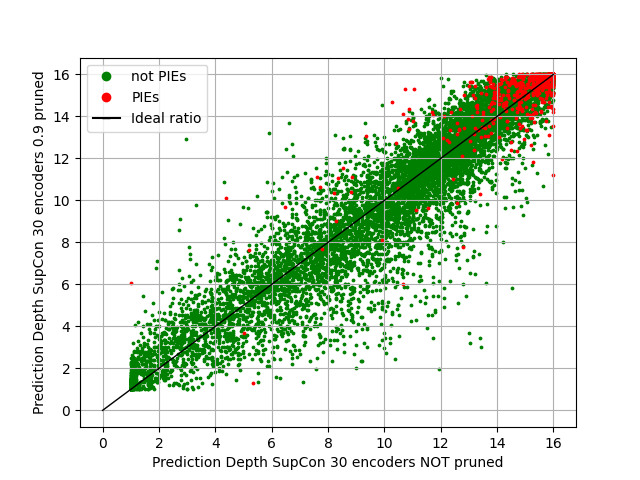}
    \caption{\textbf{\pdepth analysis of the impact of pruning on the quality of learned representations.} In all plots \pdepth of a non-pruned network (x-axis) is compared to the \pdepth of a pruned network with 90\% sparsity (y-axis). \textbf{Left to right}: \supervised~/~\gmp, \supervised~/~\oneshot, \supcon~/~\gmp, \supcon~/~\oneshot. For \supervised models, \pdepth of a sample in uncompressed model is a good predictor of \pdepth of a pruned model (points lie close to the identity line). For \supcon \pdepth of a sample may change drastically due to pruning, thus high variance in two \supcon plots for both \gmp and \oneshot pruning on the right. See Appendix \Secref{sec:scores} for further analysis.}
    \label{fig:pdepth-appendix}
\end{figure*}

\paragraph{\qscore.}
    Recently introduced by \citet{kalibhat2022understanding}, \qscore is an unsupervised metric to quantify the quality of the latent representation of a sample produced by an encoder network. It can be used as a regularizer during training to let the network produce a better latent representation. As shown in \citep{kalibhat2022understanding}, learned representations of correctly classified samples (1) have few discriminative features, \ie higher feature sparsity, and (2) the feature values deviate significantly from the values of other features. These two properties are encoded in \qscore defined as follows:
\begin{equation}
    \text{Q}_{i} = \frac{\text{Z-Score}(h_{i})}{\| h_{i} \|_{1}},
\end{equation}
\noindent where  $h_{i}$ is a latent representation of a sample $i$, $\| h_{i} \|_{1}$ is its L1 norm and Z-Score is computed as $\frac{\mathrm{max}(h_{i} - \mu_{i})}{\sigma_{i}}$. A sample with a high \qscore has high probability to be classified correctly~\citep{kalibhat2022understanding}.

\paragraph{Prediction depth (\pdepth).}
\pdepth introduced in \citep{baldock2021deep} is a supervised metric which measures sample difficulty based on the depth of a network required to correctly represent the sample. It uses an encoder network where, for each representation, a $k$-Nearest Neighbors (kNN)~\citep{fix1952discriminatory, cover1967nearest} classifier is trained using hidden representations of each sample in the training set. The prediction depth score of a sample is determined by the depth from which its hidden representation is correctly classified for all the subsequent kNNs.  \citep{baldock2021deep} show that difficult samples, \ie samples that are easily misclassified, have high prediction depth. 

\paragraph{UMAP.}
In this work we also illustrate the UMAP of the  representation vectors after applying different pruning methods. UMAP \cite{mcinnes2018umap} is a dimensionality reduction algorithm created over a mathematical framework which preserves pairwise and global distance structures of the data projected onto a lower dimensional space. Compared to other dimensionality reduction algorithms, UMAP has no restrictions on the size of the embedding vectors.

\begin{figure}[t]
    \centering
    \begin{subfigure}[b]{\linewidth}
        \includegraphics[width=0.32\linewidth]{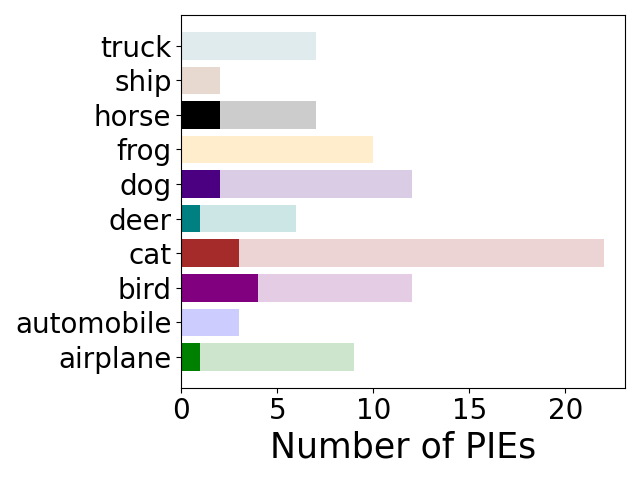}
        \includegraphics[width=0.32\linewidth]{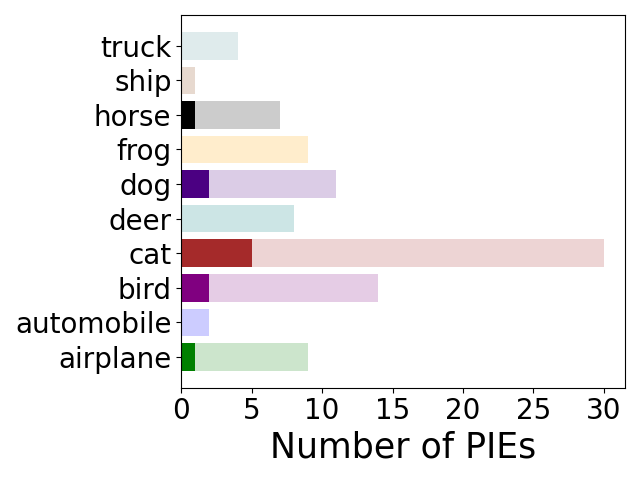}   
        \includegraphics[width=0.32\linewidth]{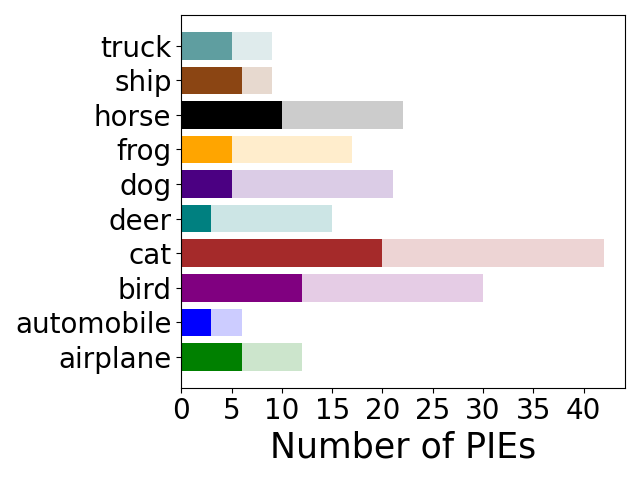} 
        \caption{PIEs obtained by \oneshot pruning \supervised models: 30\% (\textbf{left}), 50\% (\textbf{center}) and 90\% (\textbf{right}) sparsity.}
    \end{subfigure}
    \hfill
    \begin{subfigure}[b]{\linewidth}
        \includegraphics[width=0.32\linewidth]{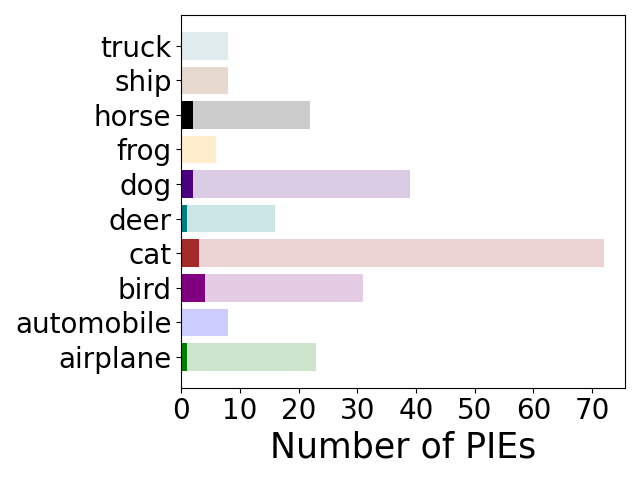}
        \includegraphics[width=0.32\linewidth]{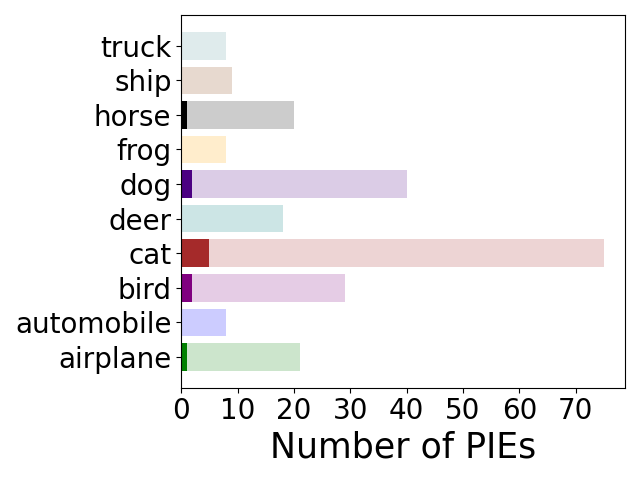}
        \includegraphics[width=0.32\linewidth]{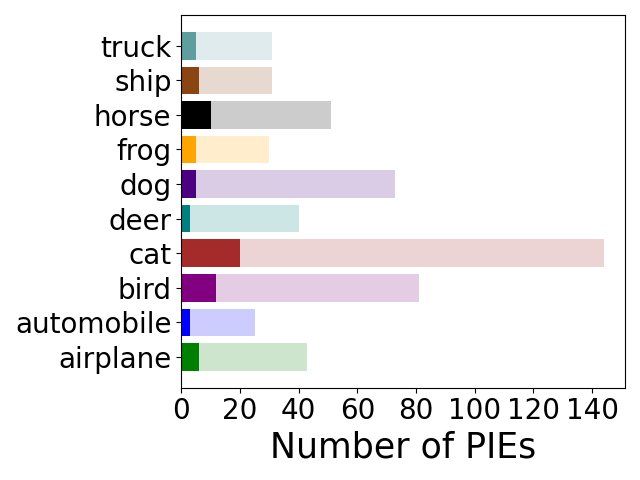} 
        \caption{PIEs obtained by \oneshot pruning \supcon models: 30\% (\textbf{left}), 50\% (\textbf{center}) and 90\% (\textbf{right}) sparsity.}
    \end{subfigure}
     \caption{\textbf{Impact of pruning on PIE distribution with 30 models.} Strong colors represent the number of PIEs shared between \supervised and \supcon models; light colors represent unique PIEs.} 
    \label{tab:pies-distribution}
\end{figure}

\section{Evaluation}
\label{sec:evaluation}

Following \citep{hooker2019compressed}, 30 WideResNet16-2 models were trained with SGD on CIFAR10 for each sparsity and learning method. The length of the representation vector $l$ is chosen equal to 128. We prune the networks to achieve sparsity levels $t \in \{0, 30, 50, 70, 90\}$\%. \gmp~\cite{zhu2017prune} and \gmpdelayed prune models during training, whereas \oneshot applies pruning post-training followed by finetuning. Below we compare the results obtained for \supervised and \supcon models with respect to the distribution of PIEs and compare the representation obtained by pruned models to the one learned by uncompressed models.

\subsection{Impact of Sparsity on Distribution of PIEs}

The model test-set accuracy and the total number of counted PIEs are summarized in \tabref{tab:PIEsComparison}. The comparison between \gmp and \oneshot pruning for \supervised models shows that the latter yields the lower number of PIEs across all pruning ratios. Among \supcon models, the lowest number of PIEs is achieved by \gmpdelayed across all sparsity levels (except for 70\% of sparsity). We also observe that \supcon models forget more easily than \supervised models for sparsities over 30\%.

The performance sensitivity with respect to individual classes is amplified as sparsity increases. In \Figref{tab:pies-distribution} we observe a similar behavior in our experiments as reported in  \citep{hooker2019compressed}: The number of samples being lost by \supervised and \supcon models gets more uniformly distributed across classes with higher sparsity. This uniform loss phenomenon is best observed when 95\% and 99\% compression is applied to the models, also shown in Appendix \Figref{fig:pie-distribution-appendix}.

\subsection{Impact of Sparsity on Representation Quality}
We compute \qscore for each sample by taking an average of 30 pruned and non-pruned models. \qscore decreases with higher sparsity and this trend is considerably more pronounced in \supcon than in \supervised models for both \gmp and \oneshot pruning algorithms, as shown in  \Figref{fig:qscore-appendix} for 90\% sparsity. The average \qscore obtained for \supcon and \supervised models for other sparsity levels is reported in  \tabref{tab:qscore} and \tabref{tab:qscore-appendix} in the appendix. Even though we observe higher \qscore for \supervised than for \supcon learning, we believe it is incorrect to compare \qscore across learning methods due to differences in the loss function and different implicit bias. Note that high variance in \qscore, \zscore and $||h_i||_1$ prevents a more fine-grained analysis.

\pdepth for each sample is computed by taking the average of 5 pruned or 5 non-pruned models. In \Figref{fig:pdepth-appendix} the \pdepth of samples processed with \supervised and \supcon models for 90\% sparsity is 
compared to \pdepth computed for uncompressed models. Average \pdepth for different sparsity levels is reported in Appendix \tabref{tab:pdepth}. \pdepth is higher for PIEs and lower for the correctly classified samples. For \supervised models, \pdepth of a sample in uncompressed model is a good predictor of \pdepth of that sample in a pruned model. For \supcon, \pdepth of a sample may change drastically due to pruning, and thus we observe high variance in two \supcon plots on the left for both \gmp and \oneshot pruning.

A qualitative analysis of the sparsified representations based on UMAP and visualization of individual PIEs is discussed in \Secref{sec:qualitative} in the Appendix. We show that difficult examples are largely unique for a training and pruning combination for low sparsity values even though the distribution of PIEs over classes looks similar. This is a peculiar phenomenon that requires further investigation.

\section{Discussion}
\label{sec:discussion}

This paper explores the impact of magnitude pruning on supervised contrastive learning and compares the impacted representation quality to supervised learning using several metrics described in the literature. Our findings suggest that models optimized with a contrastive objective are more sensitive to the introduction of sparsity relatively to traditional supervised training with a cross-entropy loss. Sparsity in these optimization settings significantly increases disparate error between classes at high levels of sparsity. In particular, sparsification of models trained using a contrastive objective leads to $\times$3 as many misclassified examples as supervised learning models sparsified with global post-training magnitude pruning, in the best setting. Curiously enough, difficult examples are largely unique across considered training methods for low sparsities, but the overlap grows with higher sparsity. We hope this work sparks interest in developing more representation-friendly pruning methods for supervised contrastive learning.


\clearpage
\bibliography{SelfPIE.bib}
\bibliographystyle{icml2022}
\clearpage
\appendix
\section*{Appendix}
\section{Training parameters}
\label{sec:training_params}

Table \ref{tab:training_hyperparameters} summarizes the set of used hyper-parameters for training different networks. We refer to the corresponding papers describing the details of the \supcon loss function and the data augmentation methods.

All experiments were performed on the TU Graz computing infrastructure.

\begin{table}[h]
\footnotesize
    	\centering
    	\begin{tabular}{cccc}
    		\toprule
    		\textbf{\textbf{Hyperparameter}} & \textbf{\supervised}& \textbf{\supcon} \\ 
    		\midrule
            Dropout          & \multicolumn{2}{c}{Not used} \\
            Weight decay     & \multicolumn{2}{c}{5e-4}     \\
            Batch size       & 128  & 1024 \\
            \# epochs        & 205  & 500 \\
            Loss function    & \small{Cross-entropy} & \small{\citep{khosla2020supervised}} \\ 
            Optimizer        & \multicolumn{2}{c}{SGD} \\
            Learning rate    & 1.0  & 0.05 \\
            Momentum         & \multicolumn{2}{c}{0.9} \\
            \multirow{2}{*}{Data Augment.} & Rnd. Crop, &                    \multirow{2}{*}{\small{\citep{arxiv.2010.08127}}}  \\
            & Horiz. Flip \\
            Label smoothing                               & \multicolumn{2}{c}{Not used}\\
            Sparsity begin step   & 1000  & 125  \\
            Sparsity end step     & 20000 & 2476 \\
            Pruning frequency     & 500   & 62   \\
            Cosine Annealing      & Not used & Used \\ 
            Warmup epochs         & \multicolumn{2}{c}{Not used} \\
            Temperature           & Not used & 0.5 \\ 
            Network type          & \multicolumn{2}{c}{WideResNet}   \\
    		\bottomrule
    	\end{tabular}
    	\caption{\textbf{Training hyper-parameters}.}
    	\label{tab:training_hyperparameters}
\end{table}

\section{PIE Comparison}

\begin{table}[t]
\centering
	\begin{tabular}{c|ccc}
    \toprule
    \multirow{2}{*}{\textbf{Sparsity}} &
    \textbf{\supervised\;/\;\gmp$^\star$} &
    \multicolumn{2}{c}{\textbf{\supervised\;/\;\gmp}} \\
    &
    \textbf{{\small{\citet{hooker2019compressed}}}} &
    \multicolumn{2}{c}{\textbf{(Ours)}} \\
    \% &
    PIE &
    PIE & Acc[\%] \\
    \midrule
    0   & -    & -   & 90.58  $\pm 0.38$   \\
    30   & 114 & 188 & 90.5 $\pm 0.39$  \\
    50   & 144 & 179 & 90.4 $\pm 0.34$ \\
    70   & 137 & 204 & 90.14 $\pm 0.34$ \\
    90   & 216 & 273 & 89.47 $\pm 0.36$ \\
    \bottomrule
    \end{tabular}
    \caption{\textbf{Comparison of the number of PIEs for different training pruning methods.} Our results differ from those reported in \cite{hooker2019compressed} due to a different model architecture used.} 
    \label{tab:PIEsComparison2}
\end{table}

We closely follow the setup by \citet{hooker2019compressed} to compute the number of PIEs for the \supervised\,/\,\gmp setting. We obtain slightly different results due to the use of a different network architecture. The comparison is detailed in \tabref{tab:PIEsComparison2}.

\section{Representation Quality Scores}
\label{sec:scores}

\begin{figure*}[t]
    \centering
    \includegraphics[width=0.245\linewidth]{img/umap/UMAP_Classes_supervised_sparsity0.0.png}
    \includegraphics[width=0.245\linewidth]{img/umap/UMAP_Classes_SupCon_sparsity0.0.png} \\
    \includegraphics[width=0.245\linewidth]{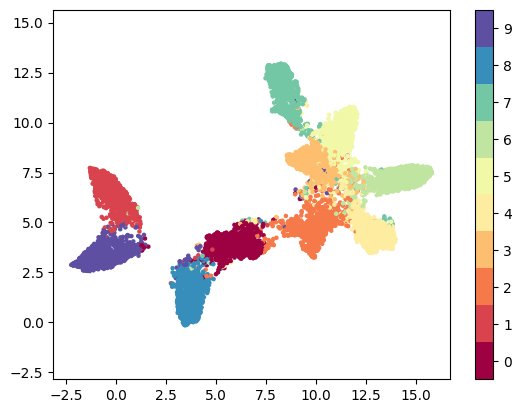}
    \includegraphics[width=0.245\linewidth]{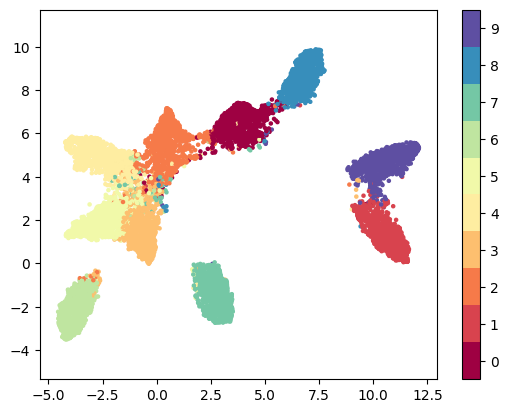}
    \includegraphics[width=0.245\linewidth]{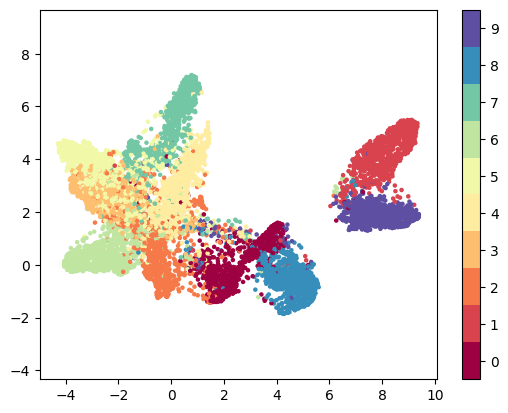}
    \includegraphics[width=0.245\linewidth]{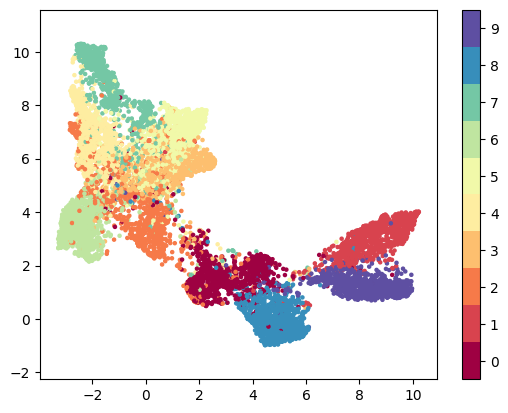} \\
    \includegraphics[width=0.245\linewidth]{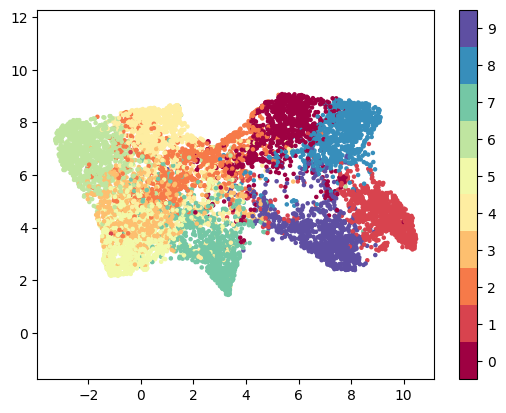}
    \includegraphics[width=0.245\linewidth]{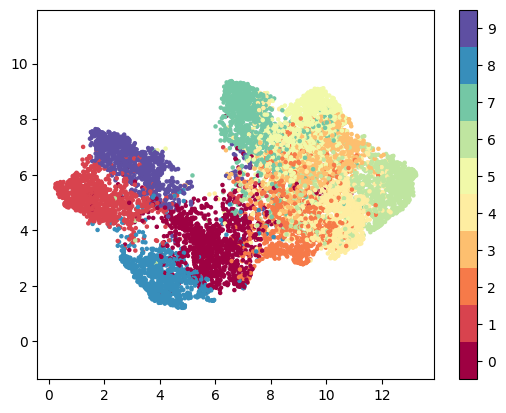}
    \includegraphics[width=0.245\linewidth]{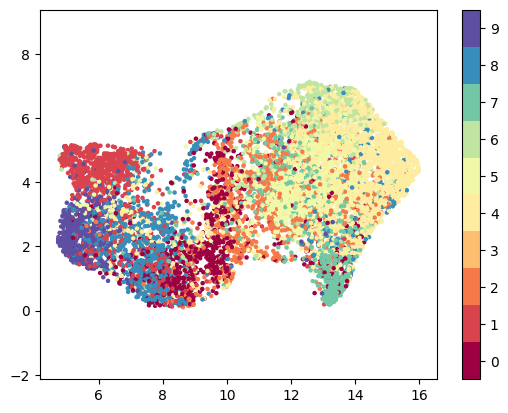}
    \includegraphics[width=0.245\linewidth]{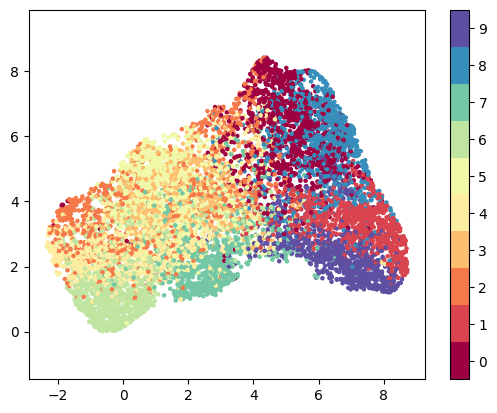} 
    \caption{\textbf{UMAP analysis of the impact of different pruning methods on the learned representations.} The top row presents the results for \supervised (\textbf{left}) and \supcon (\textbf{right}) models trained with 0\% sparsity.
    The second and the third rows present the results when applying 90\% and 99\% sparsity, respectively.
    \textbf{Left to right}: \supervised\;/\;\gmp, \supervised\;/\;\oneshot, \supcon\;/\;\gmp, \supcon\;/\;\oneshot.
    }
    \label{fig:umap-appendix}
\end{figure*}

A detailed comparison of UMAPs obtained for different sparsities is shown in \Figref{fig:umap-appendix}. We observe that representations learned by \supcon degrade fast with higher sparsity resulting in a cloud of points. Although UMAP diagrams do not reflect discriminator performance, we can observe that representation projections of \supervised and \supcon models are fairly different.

\begin{table*}[t]
\centering
	\begin{tabular}{c|cccccc}
    \toprule
    \textbf{Sparsity} &
    \multicolumn{3}{c}{\textbf{\supcon\;/\;\gmp}} & 
    \multicolumn{3}{c}{\textbf{\supcon\;/\;\oneshot}} \\
    \% &
    Q & Z & $||h_i||_1$ &
    Q & Z & $||h_i||_1$ \\
    \midrule
    0  & 0.72$\pm 0.17$ & 5.01$\pm 0.62$ & 7.15$\pm 0.8$ & 0.72$\pm 0.16 $ & 5.01$\pm 0.62$ & 7.15$\pm 0.8$ \\ 
    30 & 0.69$\pm 0.16$ & 4.95$\pm 0.65$ & 7.34$\pm 0.81$ & 0.66$\pm 0.14$ & 4.92$\pm 0.61$ & 7.6$\pm 0.76$ \\
    50 & 0.66$\pm 0.16$ & 4.84$\pm 0.64$ & 7.5$\pm 0.82$ & 0.65$\pm 0.14 $ & 4.93$\pm 0.63$ & 7.71$\pm 0.76$ \\
    70 & 0.62$\pm 0.15$ & 4.68$\pm 0.6$ & 7.74$\pm 0.82$ & 0.63$\pm 0.08$ & 4.96$\pm 0.63$ & 7.95$\pm 0.74$ \\ 
    90 & 0.56$\pm 0.13$ & 4.47$\pm 0.57$ & 8.14$\pm 0.82$ & 0.59$\pm 0.01$ & 4.89$\pm 0.57$ & 8.39$\pm 0.66$ \\ 
    \bottomrule
    \end{tabular}
    \caption{\textbf{Representation quality assessed by \qscore for \supcon trained models.} Representation quality of \supcon models is affected by pruning. This can be seen by a fast degradation of the \qscore value with higher sparsity compared to the \supervised setting (see \tabref{tab:qscore-appendix}). Sparsity affects both \zscore and $||h_i||_1$ resulting in a lower \zscore and a higher $||h_i||_1$.}
    \label{tab:qscore}
\end{table*}

\begin{table*}[t]
\centering
	\begin{tabular}{c|cccccc}
    \toprule
    \textbf{Sparsity} &
    \multicolumn{3}{c}{\textbf{\supervised\;/\;\gmp}} & 
    \multicolumn{3}{c}{\textbf{\supervised\;/\;\oneshot}} \\
    \% &
    Q & Z & $||h_i||_1$ &
    Q & Z & $||h_i||_1$ \\
    \midrule
    0  & 1.44$\pm 0.29$ & 5.73$\pm 0.68$ & 4.15$\pm 0.4$ & 1.44$\pm 0.29$ & 5.73$\pm 0.68$ & 4.15$\pm 0.4$ \\ 
    30 & 1.43$\pm 0.3$ & 5.71$\pm 0.69$ & 4.19$\pm 0.41$ & 1.41$\pm 0.28$ & 5.68$\pm 0.67$ & 4.19$\pm 0.39$ \\
    50 & 1.41$\pm 0.29$ & 5.66$\pm 0.68$ & 4.2$\pm 0.41$ & 1.42$\pm 0.29$ & 5.69$\pm 0.68$ & 4.19$\pm 0.39$ \\
    70 & 1.39$\pm 0.28$ & 5.61$\pm 0.65$ & 4.2$\pm 0.4$ & 1.42$\pm 0.26$ & 5.69$\pm 0.67$ & 4.19$\pm 0.39$ \\ 
    90 & 1.39$\pm 0.27$ & 5.66$\pm 0.64$ & 4.21$\pm 0.38$ & 1.38$\pm 0.26$ & 5.72$\pm 0.66$ & 4.3$\pm 0.35$ \\ 
    \bottomrule
    \end{tabular}
    \caption{\textbf{Representation quality assessed by \qscore for \supervised trained models.} Representation quality of \supervised models is not affected as sparsity increases, since \zscore and $||h_i||_1$ stay close to the values of non-pruned models (0\% sparsity).} 
    \label{tab:qscore-appendix}
\end{table*}

Complementing \Figref{fig:qscore-appendix}  discussed in the main paper, \tabref{tab:qscore} and \tabref{tab:qscore-appendix} show the performance of \gmp and \oneshot pruning methods for \supcon and \supervised learning respectively. Even though we observe higher values of \qscore for \supervised than for \supcon learning, we believe it is not correct to compare \qscore values across different representation learning methods, mainly due to differences in the loss function and different implicit bias. However, when comparing \gmp to \oneshot pruning for \supervised models, both perform fairly stable over a large range of sparsities. This is different for \supcon models, where \qscore shows a significant drop with higher sparsity. 

\begin{table*}[t]
\small{
\centering
	\begin{tabular}{c|cccc|cccc}
    \toprule
    \textbf{Sparsity} &
    \multicolumn{2}{c}{\textbf{\supervised\;/\;\gmp}} & 
    \multicolumn{2}{c}{\textbf{\supervised\;/\;\oneshot}} & 
    \multicolumn{2}{c}{\textbf{\supcon\;/\;\gmp}} & 
    \multicolumn{2}{c}{\textbf{\supcon\;/\;\oneshot}} \\
    \% &
    PIE & non-PIE &
    PIE & non-PIE &
    PIE & non-PIE &
    PIE & non-PIE \\
    \midrule
    0  & 14.87$\pm0.96$ & 10.85$\pm 2.82$ &  14.87$\pm 0.96$ & 10.85$\pm 2.82$ & 14.67$\pm 1.45$  & 9.94$\pm 2.91$   & 14.67$\pm 1.45$ & 9.94$\pm 2.91$ \\
    30 & 14.87$\pm 1.01$ & 10.98$\pm 2.75$ & 14.95$\pm 1.09$ & 11$\pm 2.82$  & 14.95$\pm 1.91$ & 10.35$\pm 4.15$ & 14.67$\pm 1.86$ & 9.67$\pm 4.03$ \\
    50 & 14.89$\pm 1.1$ & 10.84$\pm 2.94$ & 14.92$\pm 1.17$ & 10.97$\pm 2.93$ & 15.09$\pm 1.64$ & 10.36$\pm 4.2$ & 14.69$\pm 1.99$ & 9.71$\pm 4.06$ \\
    70 & 15.03$\pm$ 1.2& 11.01$\pm 2.86$ & 14.85$\pm 1.18$ & 10.89$\pm 2.86$ & 15.14$\pm 1.81$ & 10.44$\pm 4.23$ & 14.56$\pm 2.23$ & 9.79$\pm 4.03$ \\
    90 & 15.1$\pm 1.13$ & 11.03$\pm 2.91$ &  14.97$\pm 1.13$ & 10.89$\pm 2.91$ & 15.15$\pm 1.71$ & 10.48$\pm 4.29$ & 14.97$\pm 1.7$ & 10$\pm 4.18$ \\
    \bottomrule
    \end{tabular}
    \caption{\textbf{Representation quality assessed by \pdepth for \supervised and \supcon trained models.} \pdepth of \supcon models has higher variance than for \supervised models indicating that pruning is oblivious to representation quality learned by \supcon.}
    \label{tab:pdepth}
}
\end{table*}

\pdepth analysis for 90\% sparsity is visualized in \Figref{fig:pdepth-appendix}. As can be observed from the plots obtained for \supervised learning the \pdepth of sparse models and uncompressed models tend to be equal, points lie close to the ideal ratio line. For \supcon models instead \pdepth points are more scattered, \pdepth of sparse \supcon models is not similar to \pdepth obtained from uncompressed \supcon models. 
The same effect can also be observed in \tabref{tab:pdepth}: The variance of \pdepth for \supcon models is higher than for \supervised models.
These observations confirm that \supcon models' behavior is significantly different  when pruning is applied, while for \supervised models this is not the case.

\section{PIE Distribution and Visualization}
\label{sec:qualitative}

\begin{figure*}
    \centering
    \begin{subfigure}{.22\textwidth}
      \centering
      \includegraphics[width=0.9\linewidth]{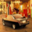}
      \caption*{U: Ship S: Automobile}
    \end{subfigure}%
    \begin{subfigure}{.22\textwidth}
      \centering
      \includegraphics[width=0.9\linewidth]{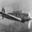}
      \caption*{U: Ship S: Airplane}
    \end{subfigure}
    \begin{subfigure}{.22\textwidth}
      \centering
      \includegraphics[width=0.9\linewidth]{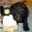}
      \caption*{U: Cat S: Dog}
    \end{subfigure}
    \begin{subfigure}{.22\textwidth}
      \centering
      \includegraphics[width=0.9\linewidth]{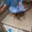}
      \caption*{U: Airplane S: Cat}
    \end{subfigure}
    \begin{subfigure}{.22\textwidth}
      \centering
      \includegraphics[width=0.9\linewidth]{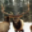}
      \caption*{U: Cat S: Deer}
    \end{subfigure}%
    \begin{subfigure}{.22\textwidth}
      \centering
      \includegraphics[width=0.9\linewidth]{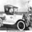}
      \caption*{U: Truck S: Automobile}
    \end{subfigure}
    \begin{subfigure}{.22\textwidth}
      \centering
      \includegraphics[width=0.9\linewidth]{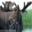}
      \caption*{U: Ship S: Deer}
    \end{subfigure}
    \begin{subfigure}{.22\textwidth}
      \centering
      \includegraphics[width=0.9\linewidth]{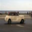}
      \caption*{U: Airplane S: Automobile}
    \end{subfigure}
    \begin{subfigure}{.22\textwidth}
      \centering
      \includegraphics[width=0.9\linewidth]{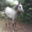}
      \caption*{U: Deer S: Horse}
    \end{subfigure}%
    \begin{subfigure}{.22\textwidth}
      \centering
      \includegraphics[width=0.9\linewidth]{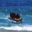}
      \caption*{U: Deer S: Ship}
    \end{subfigure}
    \begin{subfigure}{.22\textwidth}
      \centering
      \includegraphics[width=0.9\linewidth]{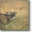}
      \caption*{U: Airplane S: Deer}
    \end{subfigure}
    \begin{subfigure}{.22\textwidth}
      \centering
      \includegraphics[width=0.9\linewidth]{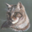}
      \caption*{U: Dog S: Cat}
    \end{subfigure}

    \caption{Correctly classified samples by \supcon 30\% sparse model and missclassified by \supcon dense model. The eight images on top (first two rows) are correctly classified by \gmp pruned model while the four images on bottom are correctly classified by \oneshot pruned model. For each image, the class prediction from the uncompressed model (U) is shown on the left while for the sparse model (S) on the right.}
    \label{fig:pie-examples-appendix}
\end{figure*}

In \Figref{fig:pie-examples-appendix} we visualize a few sample images and show the corresponding predicted classes for correctly classified images by a sparse \supcon model which are, however, wrongly classified by an uncompressed \supcon model.

\begin{figure*}[t]
    \centering
    \includegraphics[width=0.245\linewidth]{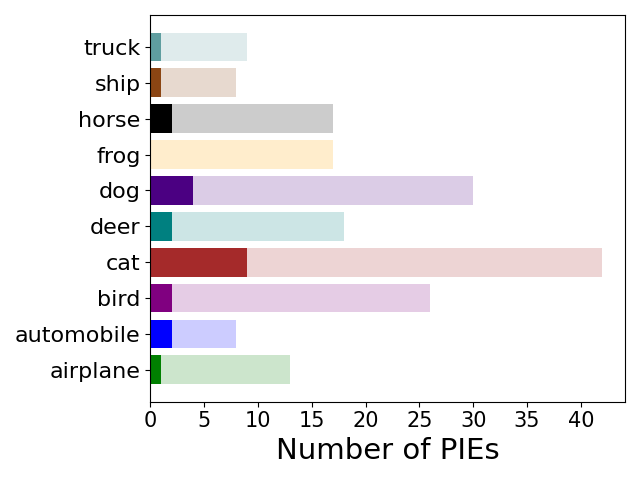}
    \includegraphics[width=0.245\linewidth]{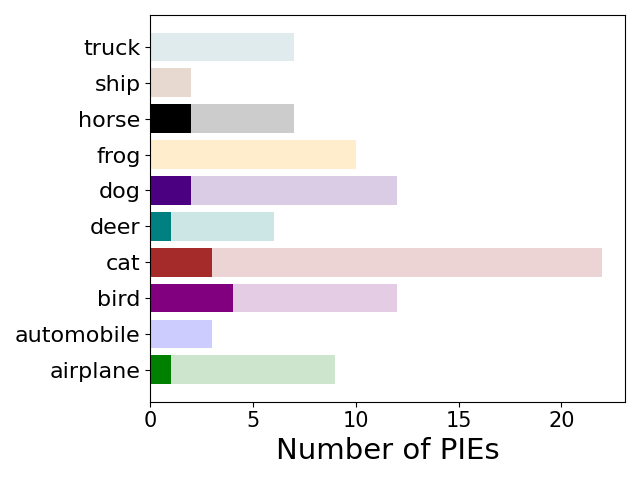}
    \includegraphics[width=0.245\linewidth]{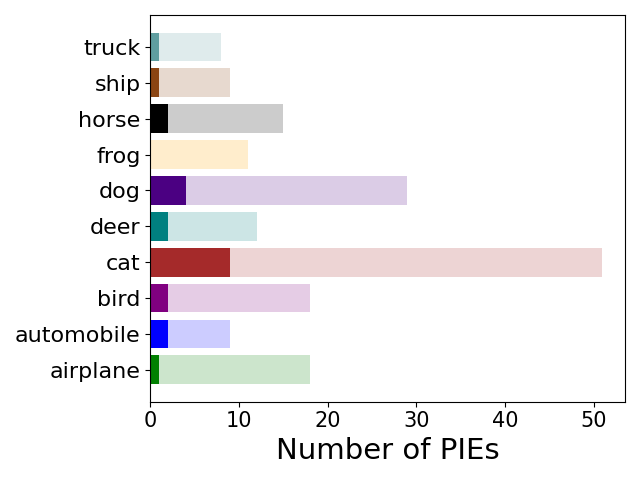}
    \includegraphics[width=0.245\linewidth]{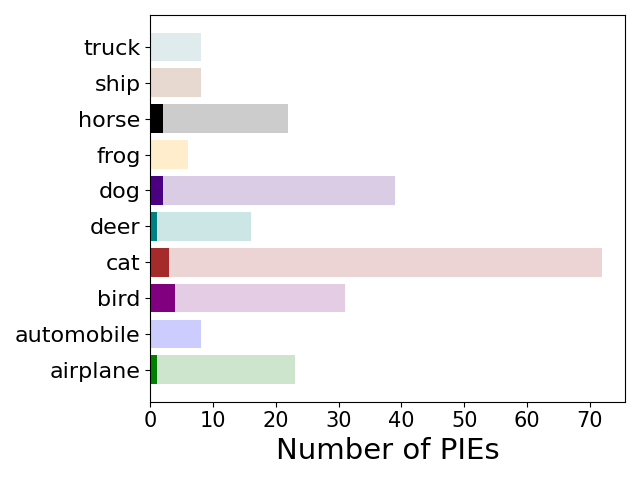} \\
    \includegraphics[width=0.245\linewidth]{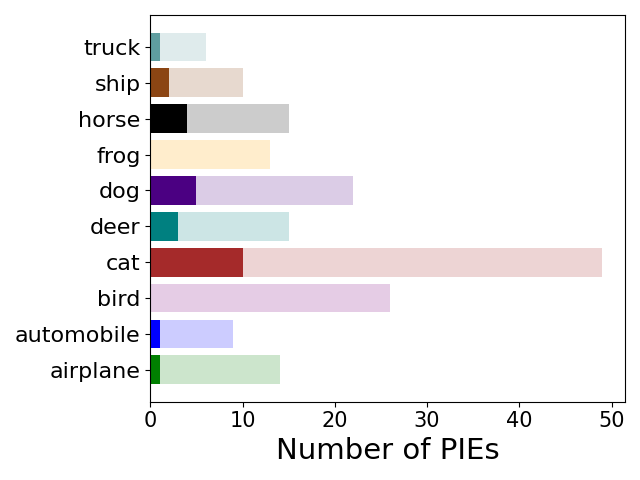}
    \includegraphics[width=0.245\linewidth]{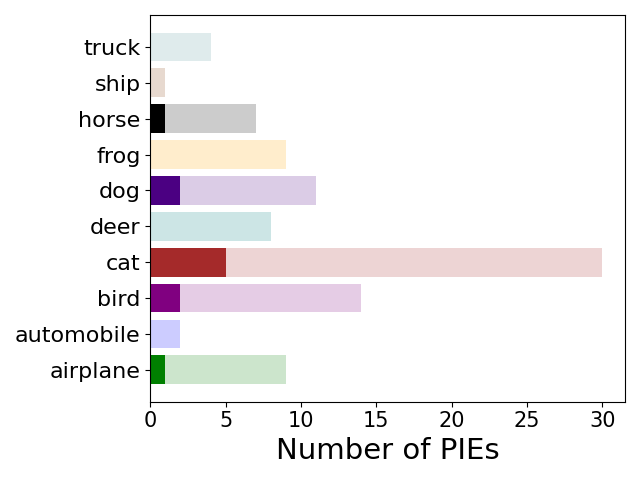}
    \includegraphics[width=0.245\linewidth]{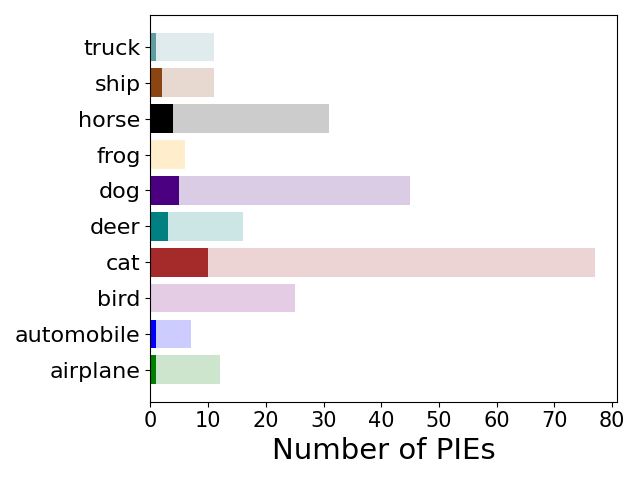}
    \includegraphics[width=0.245\linewidth]{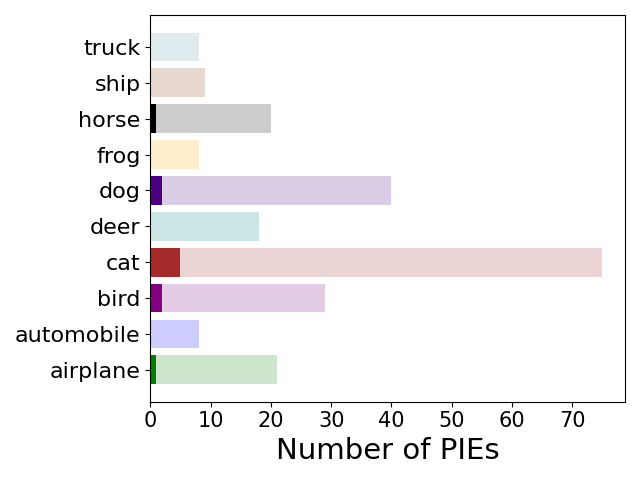} \\
    \includegraphics[width=0.245\linewidth]{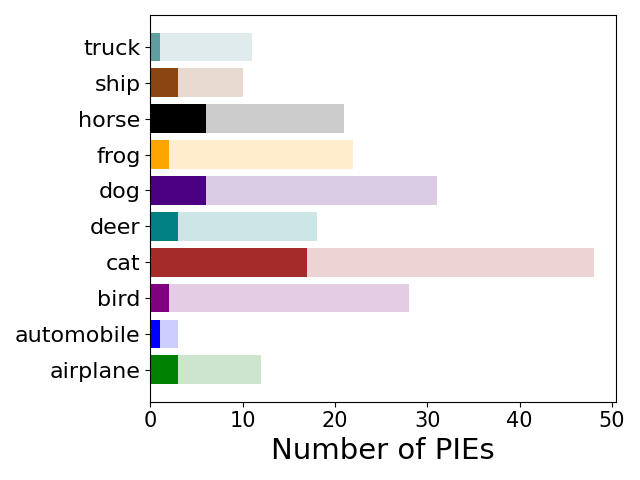}
    \includegraphics[width=0.245\linewidth]{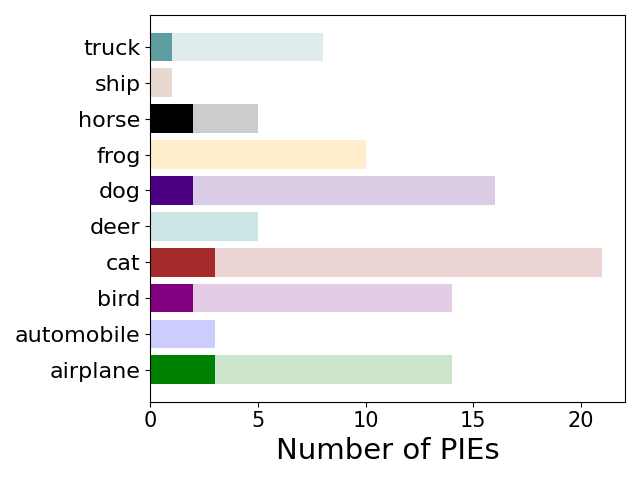}
    \includegraphics[width=0.245\linewidth]{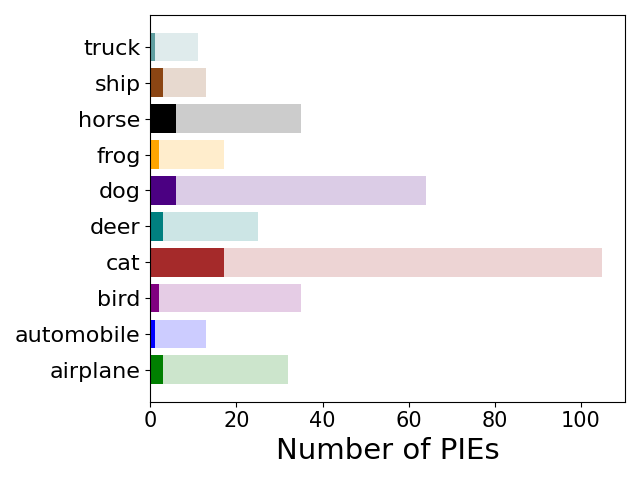}
    \includegraphics[width=0.245\linewidth]{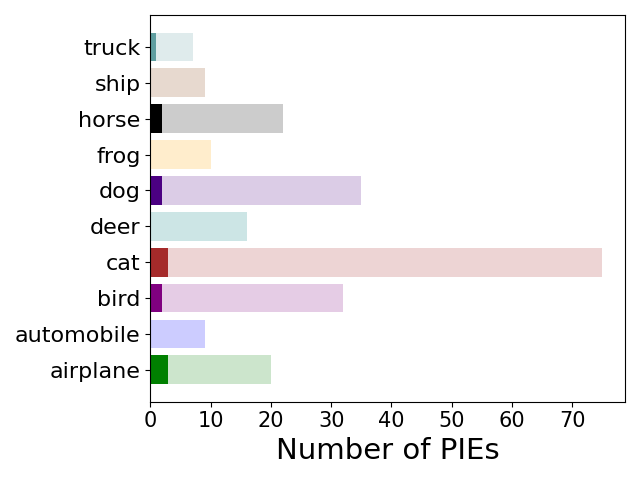} \\
    \includegraphics[width=0.245\linewidth]{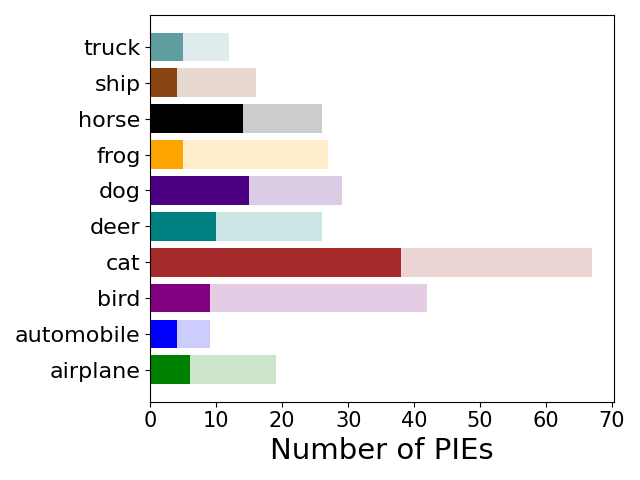}
     \includegraphics[width=0.245\linewidth]{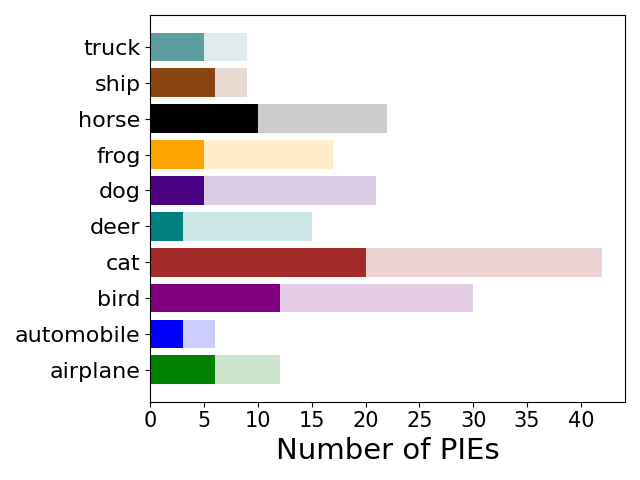}
     \includegraphics[width=0.245\linewidth]{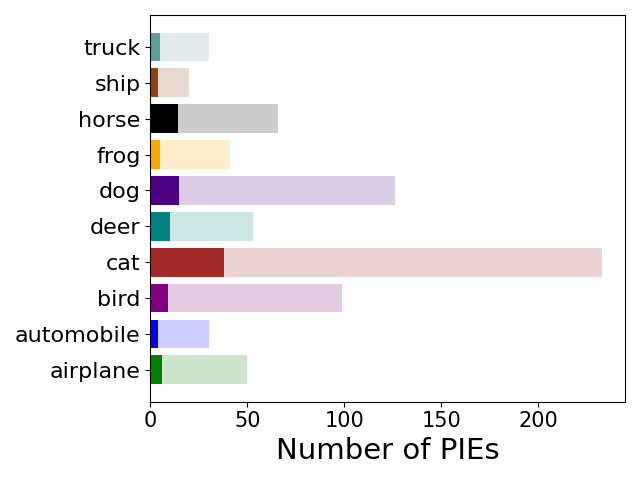}
     \includegraphics[width=0.245\linewidth]{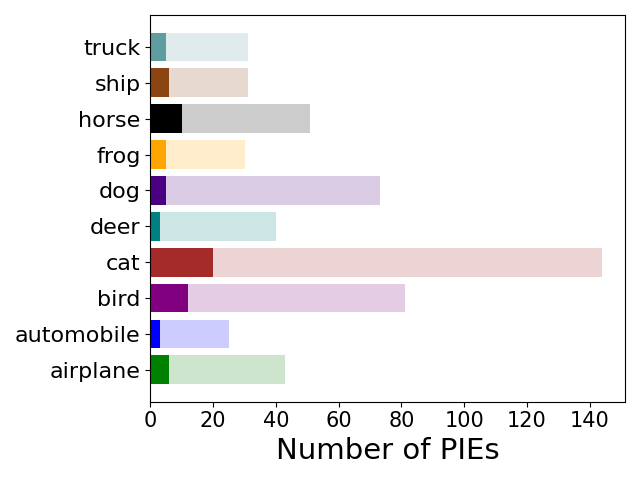} \\
     \includegraphics[width=0.245\linewidth]{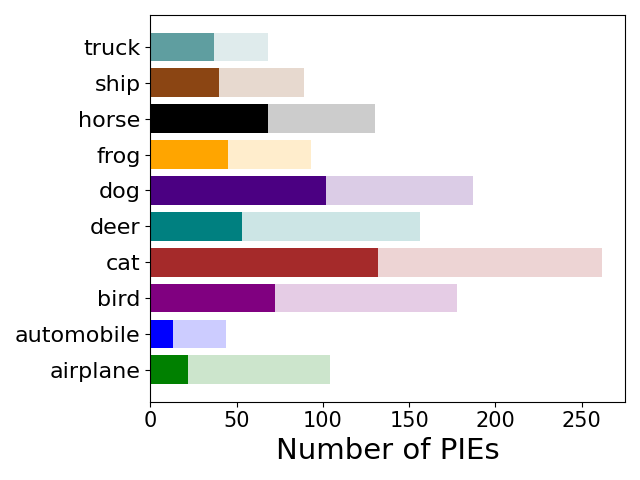}
     \includegraphics[width=0.245\linewidth]{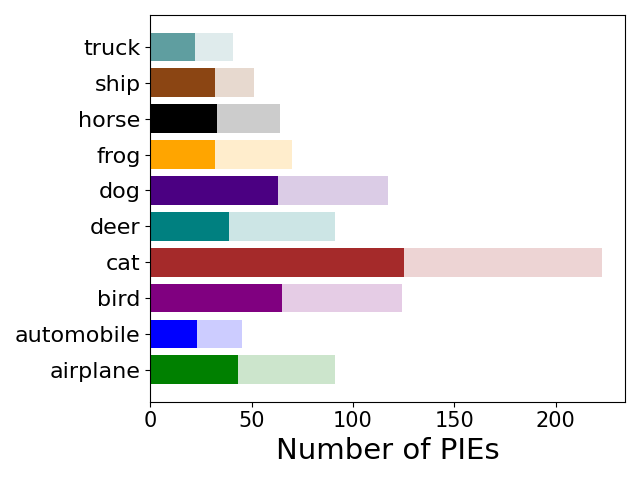}
     \includegraphics[width=0.245\linewidth]{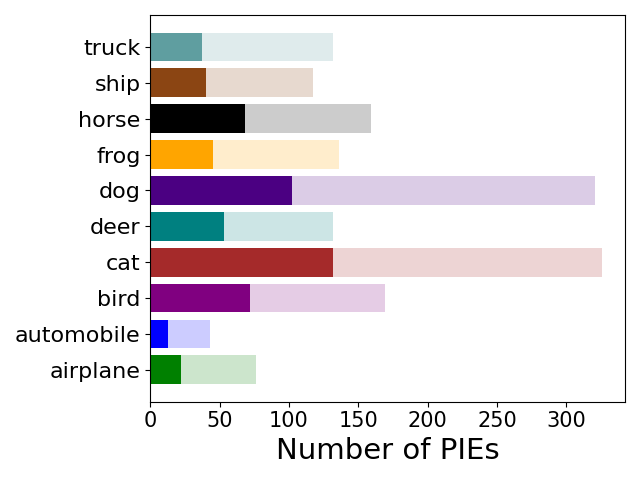}
     \includegraphics[width=0.245\linewidth]{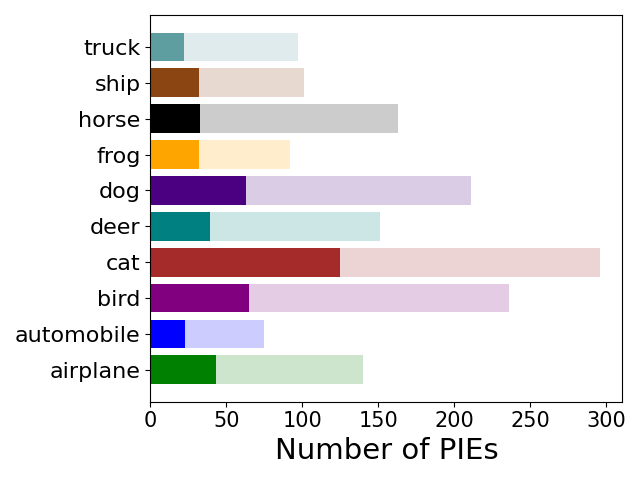} \\
     \includegraphics[width=0.245\linewidth]{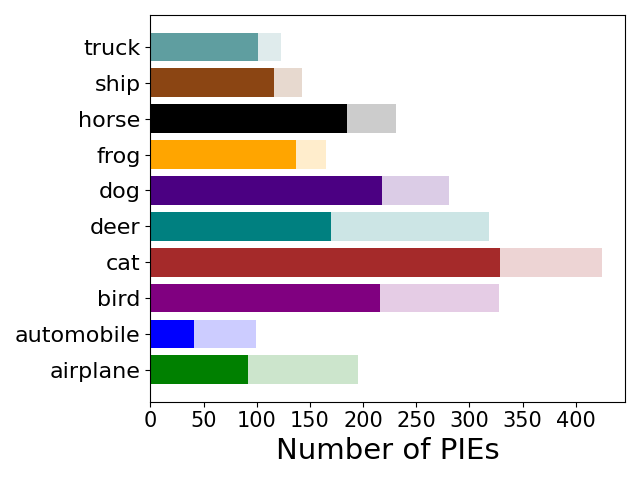}
     \includegraphics[width=0.245\linewidth]{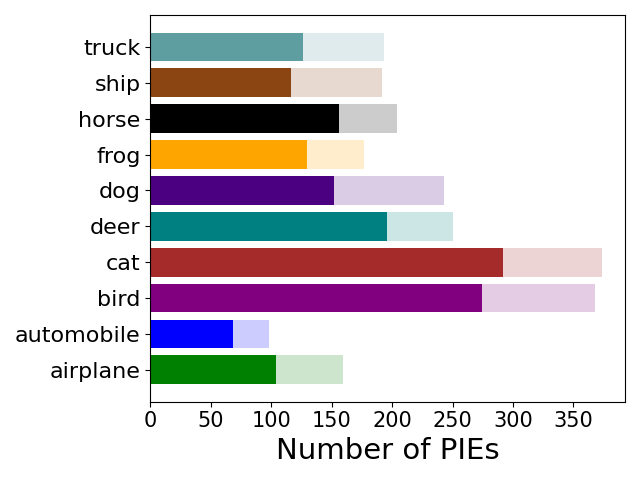}
     \includegraphics[width=0.245\linewidth]{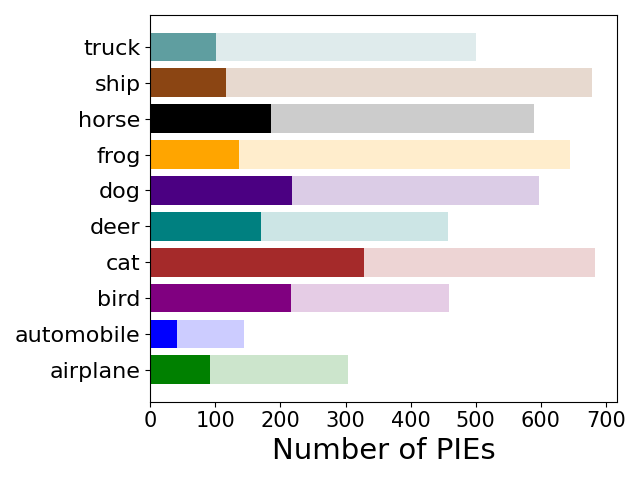}
     \includegraphics[width=0.245\linewidth]{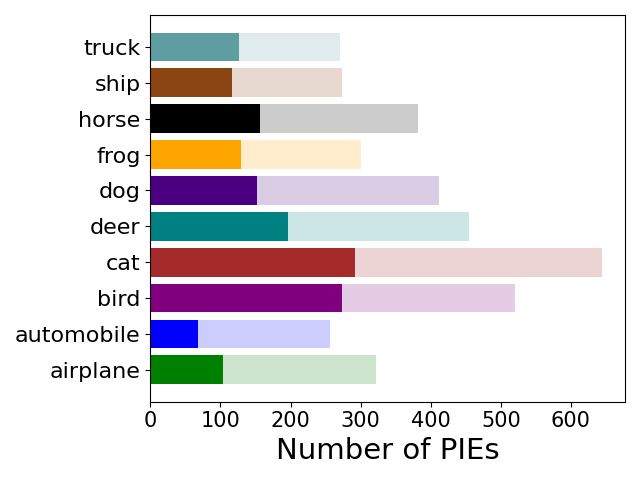}
    \caption{\textbf{PIE distribution over classes for different sparsity levels.} From \textbf{top to bottom}: 30\%, 50\%, 70\%, 90\%, 95\% and 99\% sparsity. In each column from \textbf{left to right}: \supervised~/~\gmp, \supervised~/~\oneshot, \supcon~/~\gmp, \supcon~/~\oneshot. Color bars depict the number of PIEs identified in each class. Strong colors show the number of PIEs that are the same for \supervised and \supcon models and can be considered as difficult examples regardless of the training method. Light colors show the number of unique PIEs for a training method. Curiously, the number of PIEs that are universal for both training methods is low for low sparsities, even though the distribution of PIEs over classes looks similar.} 
    \label{fig:pie-distribution-appendix}
\end{figure*}

In \Figref{fig:pie-distribution-appendix} the distribution of PIEs are visualized for all the pruning percentages. As sparsity increases PIEs distribution tends to be more uniform: For 95\% and 99\% sparsity the number of PIEs shared between \supervised and \supcon models is higher. This trend was also observed by \citep{hooker2019compressed}.

%

\end{document}